\documentclass[11pt]{article}

\usepackage{graphicx,ifthen,color,algorithm}
\usepackage{amsmath,amsthm,amssymb,amsfonts,verbatim}
\usepackage{mathrsfs,accents,setspace,eucal}
\usepackage{algorithmic,subcaption,xcolor}

\def\simind{\stackrel{{\tiny \mbox{ind.}}}{\sim}}
\def\btheta{\boldsymbol{\theta}}

\def\bbeta{\boldsymbol{\beta}}
\def\bSigma{\boldsymbol{\Sigma}}
\def\bCov{\textbf{Cov}}
\def\smhalf{{\textstyle{\frac{1}{2}}}}
\def\sigsqeps{\sigma_{\varepsilon}^{2}}

\def\bmu{\boldsymbol{\mu}}
\def\Ssc{{\mathcal S}}
\def\thickarrow{\longleftarrow}
\def\bomega{\boldsymbol{\omega}}
\def\bOmega{\boldsymbol{\Omega}}
\def\bzero{\textbf{0}}

\def\by{\textbf{\textit{y}}}
\def\bu{\textbf{\textit{u}}}
\def\bx{\textbf{\textit{x}}}
\def\bo{\textbf{\textit{o}}}
\def\bv{\textbf{\textit{v}}}
\def\bb{\boldsymbol{b}}
\def\bc{\boldsymbol{c}}
\def\bA{\textbf{\textit{A}}}
\def\bB{\textbf{\textit{B}}}
\def\bC{\textbf{\textit{C}}}
\def\bD{\textbf{\textit{D}}}

\def\bI{\textbf{\textit{I}}}

\def\bO{\textbf{\textit{O}}}

\def\bQ{\textbf{\textit{Q}}}
\def\bR{\textbf{\textit{R}}}

\def\bX{\textbf{\textit{X}}}
\def\bY{\textbf{\textit{Y}}}
\def\bZ{\textbf{\textit{Z}}}
\def\bveco{\boldsymbol{b}_1}
\def\bvect{\boldsymbol{b}_2}
\def\bveci{\boldsymbol{b}_i}
\def\bvecm{\boldsymbol{b}_m}

\def\Bmato{\bB_1}
\def\Bmatt{\bB_2}
\def\Bmati{\bB_i}
\def\Bmatm{\bB_m}

\def\Bmatdoto{\bBdot_1}
\def\Bmatdott{\bBdot_2}
\def\Bmatdoti{\bBdot_i}
\def\Bmatdotm{\bBdot_m}

\def\AtLev{\bA}
\def\SolveTwoLevelSparseLeastSquares{\textsc{\footnotesize SolveTwoLevelSparseLeastSquares}}
\def\LargerSolveTwoLevelSparseLeastSquares{\textsc{\normalsize SolveTwoLevelSparseLeastSquares}}
\def\CPU{\texttt{\footnotesize GPyT-CPU}}
\def\GPU{\texttt{\footnotesize GPyT-GPU}}
\def\SEB{\texttt{\footnotesize sEB}}
\def\smalldot{\mbox{\fontsize{0.1mm}{0.5em}\selectfont{$\bullet$}}}
\def\bBdot{\overset{\ \smalldot}{\bB}}
\def\Bmati{\bB_i}
\def\Bmatdoti{\bBdot_i}
\def\bveci{\boldsymbol{b}_i}
\def\xveco{\bx_1}
\def\xvectCi{\bx_{2,i}}
\def\AUoo{\bA^{11}}
\def\AUttCi{\bA^{22,i}}
\def\AUotCi{\bA^{12,i}}
\def\AUotCo{\bA^{12,1}} \def\AUotCoT{\bA^{12,1\,T}}
\def\AUotCt{\bA^{12,2}} \def\AUotCtT{\bA^{12,2\,T}}
\def\AUotCi{\bA^{12,i}} 
\def\AUotCm{\bA^{12,m}} \def\AUotCmT{\bA^{12,m\,T}}
\def\AUttCo{\bA^{22,1}}
\def\AUttCt{\bA^{22,2}}
\def\AUttCi{\bA^{22,i}}
\def\AUttCm{\bA^{22,m}}
\def\bigX{{\LARGE\mbox{$\times$}}}
\def\xvectCi{\bx_{2,i}}
\def\Cmatoi{\bC_{1i}}
\def\cveczi{\bc_{0i}}
\def\cvecoi{\bc_{1i}}
\def\cvecti{\bc_{2i}}
\def\Cmatzi{\bC_{0i}}
\def\Cmatoi{\bC_{1i}}
\def\Cmatti{\bC_{2i}}
\def\nadj{{\tilde n}}
\def\stackdum{\mathop{\mbox{\rm stack}}}
\def\stack#1{\stackdum_{#1}}

\def\OmegaAtwoTwo{\bOmega_4}
\begin{document}

\thispagestyle{empty}

\centerline{\Large\bf Streamlined Empirical Bayes Fitting of Linear}
\vskip2mm
\centerline{\Large\bf Mixed Models in Mobile Health}
\vskip7mm
\centerline{\normalsize\sc By Marianne Menictas$\null^1$, Sabina Tomkins$\null^2$ and
Susan A. Murphy$\null^1$}
\vskip5mm
\centerline{\textit{Harvard University$\null^1$ \quad Stanford University$\null^2$}}
\vskip6mm
\centerline{28 March, 2020}

\vskip5mm
\vskip5mm

\begin{abstract}
  To effect behavior change a successful algorithm must make high-quality decisions
  in real-time. For example, a mobile health (mHealth) application designed to increase
  physical activity must make contextually relevant suggestions to motivate users.
  While machine learning offers solutions for certain stylized settings,
  such as when batch data can be processed offline, there is a dearth of
  approaches which can deliver high-quality solutions under the specific constraints of
  mHealth. We propose an algorithm which provides users with contextualized \textit{and} personalized
  physical activity suggestions.
  This algorithm is able to overcome a challenge critical to mHealth
  that complex models be trained efficiently.
   We propose a tractable streamlined empirical Bayes procedure
  which fits  linear mixed effects models in large-data settings.
   Our procedure takes
  advantage of sparsity introduced by hierarchical random effects to efficiently
  learn the posterior distribution of a linear mixed effects model.
  A key contribution of this work is that we provide  explicit
   updates in order to learn both fixed effects, random effects and
   hyper-parameter values.
  We demonstrate the success of this approach in a
  mobile health (mHealth) reinforcement learning application, a domain in which
  fast computations are crucial for real time interventions.
  Not only is our approach computationally efficient, it is also easily
  implemented with closed form matrix algebraic updates
  and we show  improvements over state of the art  approaches
  both in speed and accuracy of up to 99\% and 56\% respectively.
\end{abstract}


\section{Introduction}

This work is motivated by a mobile health study in which an
online Thompson Sampling contextual bandit algorithm is used to personalize
the delivery of physical activity suggestions \cite{tomkins2019intelligent}.
These suggestions are intended to increase near time physical activity.
The personalization occurs via two routes, first  the user's current context
is used to decide whether to deliver a suggestion and second, random effects,
as described in Section \ref{Sec:probSetting}, are used to learn user and
time specific parameters that encode the influence  of each of the contextual
variables in this decision. The user and time specific parameters are
modeled in the reward function (the mean of the reward conditional on
context and action). To learn these parameters, information is pooled across
both users and time in a dynamic manner, combining Thompson sampling with a
Bayesian random effects model for the reward function. In contrast to fully
Bayesian methods, empirical Bayes estimates the value of hyper-parameters as
a function of observed data.

The contributions of this paper are as follows

\begin{itemize}
  \item We develop a Thompson sampling algorithm coupled
        with explicit streamlined empirical Bayes updates
        for fitting linear mixed effects models. To
        estimate hyper-parameters and compute the
        estimated posterior distribution, our algorithm computes
        closed form updates which are within the class of two-level sparse least
        squares problems introduced by \cite{nolan2019solutions}.
  \item This work provides an efficient empirical Bayes algorithm in which the amount of
        storage and computing at each iteration is $\mathcal{O}(m_1 m_2^3)$, where $m_1$ is the larger
        dimension of the two grouping mechanisms considered. For example, $m_1$ may
        represent the number of users and $m_2$ the time points  or vice-versa.
  \item Our approach reduces the running time over other state-of-art methods, and critically,
        does not require advanced hardware.
\end{itemize}

These contributions make our approach practical for online mHealth
settings, in which
incremental learning algorithm updates are required (e.g., at nightly increments),
where swift
computations are necessary for subsequent \textit{online} policy adaptation.
This facilitates incremental, accurate tuning of the variance hyper-parameters in a
Thompson-Sampling contextual bandit algorithm.

In section \ref{Sec:probSetting} we describe the problem setting and review
Thompson-Sampling with the use of a Bayesian mixed effects model for
the reward \cite{tomkins2019intelligent}.
Section \ref{Sec:PEB} describes a natural parametric empirical Bayes
approach to hyperparameter tuning and Section \ref{sec:streamPEB} presents
our streamlined alternative. A performance assessment and
comparison is shown in Section \ref{Sec:PerfAssComp}. \\

\section{Methods}

\subsection{Problem Setting}\label{Sec:probSetting}

At each time, $t$, on each user, $i$, a vector of context variables,
$\bX_{i t}$, is observed. An action, $A_{i t}$, is then selected.
Here we consider $K$ actions, where $K \in \mathbb{N}$.
Subsequently a real-valued reward, $Y_{i t}$ is observed. This continues
for $t = 1, \hdots, T$ times and on $i = 1, \hdots, m$ users.
We assume that the reward at time $t$ is generated with a person and time specific
mean, \[E[Y_{i t}| \bX_{i t}, A_{i t}]= \bZ_{i t} \bbeta + \bZ^{\bu}_{i t} \bu_{i} +
\bZ^{\bv}_{i t} \bv_{t} \] where $\bZ_{i t}= f(\bX_{i t}, A_{i t})$, $\bZ_{i t}^{\bu}=
f^{\bu}(\bX_{i t}, A_{i t})$ and $\bZ_{i t}^{\bv}= f^{\bv}(\bX_{i t}, A_{i t})$ are known
features of the context $\bX_{i t}$ and action $A_{i t}$. $(\bbeta, \bu_i, \bv_i)$ are
unknown parameters; in particular $\bu_i$ is the vector of $i$th user parameters and
$\bv_{t}$ is the vector of time $t$ parameters.  Time $t$ corresponds to
\lq\lq time-since-under-treatment\rq\rq\ for a user.
User-specific parameters, $\bu_{i}$, capture  unobserved user variables that
influence the reward at all times $t$; in mobile health unobserved user variables may
include level of social support for activity, pre-existing problems or preferences that
make activity difficult. The time-specific parameters, $\bv_{t}$ capture
unobserved \lq\lq time-since-under-treatment\rq\rq\ variables that influence the reward
for all users. In mobile health unobserved \lq\lq time-since-under-treatment\rq\rq\
variables might include treatment fatigue, decreasing motivation, etc.

The Thompson Sampling algorithm in \cite{tomkins2019intelligent} uses the following Bayesian mixed effects model 
for the reward $Y_{i t}$:
\begin{equation}
\label{MainMod1}
\begin{array}{c}
  Y_{i t} | \bbeta, \bu_{i}, \bv_{t}, \sigsqeps \simind N(\bZ_{i t}\bbeta +
  \bZ_{it}^{\bu} \bu_{i} + \bZ_{it}^{v} \bv_{t}, \ \sigsqeps).
\end{array}
\end{equation}
The algorithm is designed with independent Gaussian priors on the unknown parameters:
\begin{equation}
\label{MainMod2}
\begin{array}{c}
  \bbeta \sim N(\bmu_{\bbeta}, \bSigma_{\bbeta}), \quad
  \bu_{i} | \bSigma^{\bu} \simind N(\bzero, \bSigma^{\bu}), \ 1 \le i \le m, \\[1ex]
  \bv_{\tau} | \bSigma^{\bv} \simind N(\bzero, \bSigma^{\bv}), \ 1\le \tau \le t.
\end{array}
\end{equation}
The $\bu_i$ and $\bv_{\tau}$ are called random effects in the
statistical literature and the model in (\ref{MainMod1}) and (\ref{MainMod2})
is often referred to as a linear mixed effects model \cite{laird1982random} or a linear mixed model with crossed random effects (e.g., \cite{baayen2008mixed, jeon2017variational}).
At each time, $t$, Thompson Sampling is used to select
the action, $A_{it}$,  based on the context $\bX_{it}$. That is, we compute the
posterior distribution for $\btheta_{i t}$ where
$$
  \btheta_{i t} = [\bbeta \ \bu_{i} \ \bv_{t}]^{T},
$$
and for context $\bX_{it}=\bx$, select treatment $A_{i t} = k$
with posterior probability
\begin{equation}
\label{eqn:randProb}
  \mbox{Pr}_{
  \btheta_{it} \sim N \left( \bmu_{p(\btheta_{it})}, \bSigma_{p(\btheta_{it})}
  \right)}\Bigg( E\left[Y_{it}| \bX_{it} = \bx, A_{it} = k
  \right] = \displaystyle{\max_{a = 1, \hdots, K}} \Big\{ E[Y_{it}| \bX_{it} =
  \bx, A_{it} = a] \Big\}  \Bigg)
\end{equation}
where $\left( \bmu_{\btheta_{it}}, \bSigma_{\btheta_{it}}\right)$ are  the posterior
mean and variance covariance matrix given in the sub-blocks of (\ref{eqn:postMeanVar}).
\subsubsection{Bayesian Mixed Effects Model Components}\label{Sec:BayesMEM}
We define the following data matrices
\begin{equation*}
  \bY \equiv \left[ \bY_{1} \ \hdots \  \bY_{m} \right]^{\top} ,
  \hspace{2mm}
  \bY_{i} \equiv \left[ Y_{i1} \ \hdots \  Y_{it} \right]^{\top},
  \hspace{2mm}
  \bZ \equiv \left[ \bZ_{1} \ \hdots \  \bZ_{m} \right]^{\top} ,
  \hspace{2mm}
  \bZ_{i} \equiv \left[ \bZ_{i1} \ \hdots \  \bZ_{it} \right]^{\top},
\end{equation*}
\begin{equation*}
  \begin{array}{c}
    \bZ^{\bu}_{i} \equiv \left[
    \bZ^{\bu}_{i1} \ \hdots \ \bZ^{\bu}_{it} \right]^{\top},
      \hspace{2mm}
      \bZ^{\bv}_{i} \equiv \left[
        \begin{array}{ccc}
           \bZ^{\bv}_{i1} & \hdots & \bzero \\
           \vdots & \ddots & \vdots \\
           \bzero & \hdots & \bZ^{\bv}_{it}
        \end{array}
         \right],
      \hspace{2mm}
      \bZ^{\bu \bv} \equiv \left[
        \begin{array}{cccc}
           \bZ^{\bu}_{1} & \hdots & \bzero & \bZ^{\bv}_{1} \\
           \vdots & \ddots & \vdots  & \vdots \\
           \bzero & \hdots & \bZ^{\bu}_{m} & \bZ^{\bv}_{m}
        \end{array}
         \right],
  \end{array}
\end{equation*}
and the following parameter vectors
\begin{equation*}
  \begin{array}{c}
     \bbeta \equiv \left[ \beta_{0} \ \hdots \ \beta_{p-1}
     \right]^{\top},
     \hspace{2mm}
     \bu \equiv \left[ \bu_{1} \ \hdots \ \bu_{m} \right]^{\top},
     \hspace{2mm}
     \bv \equiv \left[ \bv_{1} \ \hdots \ \bv_{t} \right]^{\top},
  \end{array}
\end{equation*}
where, as before, $\bZ^{\bu}_{it} = f^{\bu}(\bX_{it}, A_{it})$ and
$\bZ^{\bv}_{it} = f^{\bv}(\bX_{it}, A_{it})$ represent the known features
of the context $\bX_{it}$ and action $A_{it}$.
The dimensions of matrices, for $1 \le i \le m$ and $1 \le \tau \le t$, are
\begin{equation}
  \begin{array}{c}
      \bZ_{i\tau}\ \mbox{is $1 \times p$},\ \ \bbeta\ \mbox{is $p\times1$},
      \ \ \bZ^{\bu}_{i\tau} \ \mbox{is $1 \times q_{u}$}, \bZ^{\bv}_{i\tau}\
      \mbox{is $1 \times q_{v}$},\ \ \bu_i\ \mbox{is $q_{u}\times 1$},
      \\[1ex]
      \bv_{\tau}\ \mbox{is $q_{v}\times 1$}, \ \
      \bSigma_{\bu}\ \ \mbox{is $q_{u}\times q_{u}$} \ \mbox{and}\
      \bSigma_{\bv}\ \ \mbox{is $q_{v}\times q_{v}$}.
    \label{eq:twoLevMatDimsII}
  \end{array}
\end{equation}
\subsubsection{Posterior Updates}\label{Sec:PostUpdates}
The posterior distribution $\btheta_{it}$
for the immediate treatment effect for user $i$ at time $t$ is updated
and then used to assign treatment in the subsequent time point,
$t + 1$. Here, we show the form of
the full posterior for $\left[ \bbeta \ \bu \ \bv \right]^{\top}$.
Define
\begin{equation*}
  \begin{array}{c}
    \bC \equiv \left[ \bZ \, \bZ^{\bu \bv} \right], \quad
    \bD \equiv \left[
    \begin{array}{ccc}
       \bSigma_{\bbeta}^{-1} & \bzero & \bzero \\
       \bzero & \bI_{m} \otimes \hat{\bSigma}_{\bu}^{-1} & \bzero \\
       \bzero & \bzero & \bI_{t} \otimes \hat{\bSigma}_{\bv}^{-1}
    \end{array} \right],
\end{array}
\end{equation*}
\begin{equation*}
  \begin{array}{c}
    \bR \equiv \hat{\sigsqeps} \bI, \quad \mbox{and} \quad
    \bo \equiv \left[
    \begin{array}{c}
       \bSigma_{\bbeta}^{-1} \bmu_{\bbeta} \\
       \bzero
    \end{array} \right].
  \end{array}
\end{equation*}
The estimated posterior distribution for the fixed and random reward
effects vector $\btheta$ is
\begin{equation}
  \label{eqn:FullPost}
  \begin{array}{lcl}
    \btheta \, | \, \hat{\bSigma} & \sim & N \left(
    \bmu_{p(\btheta)}, \, \bSigma_{p(\btheta)} \right),
  \end{array}
\end{equation}
where
\begin{equation}
  \label{Eqn:Sigma}
  \begin{array}{ll}
  \hat{\bSigma} \equiv (\hat{\sigsqeps}, \hat{\bSigma}_{\bu}, \hat{\bSigma}_{\bv}),
   \end{array}
\end{equation}
\begin{equation}
  \label{eqn:postMeanVar}
  \begin{array}{c}
    \bSigma_{p(\btheta)} = \left(\bC^{\top} \bR^{-1} \bC + \bD \right)^{-1},
    \quad \mbox{and}
    \quad
    \bmu_{p(\btheta)} = (\bC^{\top} \bR^{-1} \bC + \bD )^{-1}
    (\bC^{\top} \bR^{-1} \by + \bo).
  \end{array}
\end{equation}
The focus of this work is to enable fast incremental estimation of the variance
components $\bSigma \equiv (\sigsqeps, \bSigma_{\bu}, \bSigma_{\bv})$.
We describe a natural, but computationally challenging approach for estimating
these variances in Section \ref{Sec:PEB} and our
streamlined alternative approach in Section \ref{sec:streamPEB}.
\subsection{Parametric Empirical Bayes}\label{Sec:PEB}
At each time, $t$, the empirical Bayes \cite{morris1983parametric,
casella1985introduction} procedure maximizes the  marginal likelihood based on
all user data up to and including data at time $t$ with respect to $\bSigma$.
The marginal likelihood of $\bY$ is
\begin{equation*}
    \bY \, | \, \bSigma \sim N (\bzero, \, \bC \bD \bC^{\top} + \sigsqeps \bI),
\end{equation*}
and has the following form
\begin{equation*}
  \begin{array}{l}
     p(\bY \, | \, \bSigma) = (2 \pi)^{-\smhalf mt}
     |\bC \bD \bC^{\top} + \sigsqeps \bI|^{-\smhalf}
     \exp \left\{ -\smhalf \bY^{\top}
     \left( \bC \bD \bC^{\top} + \sigsqeps \bI \right)^{-1} \bY \right\}.
  \end{array}
\end{equation*}
The maximization is commonly done via  the Expectation Maximisation (EM) algorithm
\cite{dempster1977maximum}.
\subsubsection{EM Method}
The {expected} complete data log likelihood is given by
\begin{equation*}
  \begin{array}{lcl}
    L(\bSigma) & = & E\left[ \log p(\bY | \btheta, \bSigma)
     + \log p(\btheta | \bSigma) \right]
  \end{array}
\end{equation*}
where the expectation is over the distribution of $\btheta=\left[ \bbeta
\ \bu \ \bv \right]^{\top}$ given in (\ref{MainMod2}). The M-step yields
the following closed form $(\ell + 1)$ iteration estimates for the variance
components in $\hat{\bSigma}^{(\ell+1)}$:

\begin{equation}
  \label{Eqn:EM_var_updates}
  \begin{array}{lcl}
    \left(\hat{\sigma}^{2}_{\varepsilon}\right)^{(\ell + 1)} & = &
    \displaystyle{\sum_{i=1}^{m}} \displaystyle{\sum_{\tau=1}^{t}} \left\{ ||Y_{i\tau}
    - \bZ_{i\tau} \bmu_{p(\bbeta)} - \bZ^{\bu}_{i\tau} \bmu_{p(\bu_{i})}
    - \bZ^{\bv}_{i\tau} \bmu_{p(\bv_{\tau})} ||^{2} + \mbox{tr} \left(
    \bZ_{i\tau}^{\top} \bZ_{i\tau} \bSigma_{p(\bbeta)} \right) \right.
    \\[1ex]
    & & \quad\quad\quad\quad \left. + \ \mbox{tr} \left( {\bZ^{\bu}_{i\tau}}^{\top} \bZ^{\bu}_{i\tau}
    \bSigma_{p(\bu_{i})} \right) + \ \mbox{tr} \left( {\bZ^{\bv}_{i\tau}}^{\top}
    \bZ^{\bv}_{i\tau} \bSigma_{p(\bv_{\tau})} \right) \right.
    \\[2ex]
    & & \quad\quad\quad\quad \left. + \ \mbox{tr} \left( \bZ_{i\tau}^{\top} \bZ^{\bu}_{i\tau}
    \bCov_{p(\bbeta, \bu_{i})} \right) + \ \mbox{tr} \left( \bZ_{i\tau}^{\top}
    \bZ^{\bv}_{i\tau} \bCov_{p(\bbeta, \bv_{\tau})} \right) \right.
    \\[2ex]
    & & \quad\quad\quad\quad \left. + \ \mbox{tr} \left( {\bZ^{\bu}_{i\tau}}^{\top}
    \bZ^{\bv}_{i\tau} \bCov_{p(\bu_{i}, \bv_{\tau})} \right) \right\},
    \\[2ex]
    \hat{\bSigma}_{\bu}^{(\ell + 1)} & = & \frac{1}{m} \displaystyle{\sum_{i=1}^{m}} \left\{
    \bmu_{p(\bu_{i})} \bmu_{p(\bu_{i})}^{\top} + \bSigma_{p(\bu_{i})} \right\},
    \\[2ex]
    \hat{\bSigma}_{\bv}^{(\ell + 1)} & = & \frac{1}{t} \displaystyle{\sum_{\tau=1}^{t}} \left\{
    \bmu_{p(\bv_{\tau})} \bmu_{p(\bv_{\tau})}^{\top} + \bSigma_{p(\bv_{\tau})} \right\}.
  \end{array}
\end{equation}
where the posterior mean reward components for the fixed and random effects
\begin{equation}
  \label{eqn:betasubentries}
  \begin{array}{c}
      \bmu_{p(\bbeta)}, \ \bmu_{p(\bu_{i})}, \ 1 \le i \le m, \quad
      \bmu_{p(\bv_{\tau})}, \ 1 \le \tau \le t,
  \end{array}
\end{equation}
and the posterior variance-covariance reward components for the fixed and
random effects
\begin{equation}
  \label{eqn:sigmasubblocks}
  \begin{array}{c}
    \bSigma_{p(\bbeta)}, \ \bSigma_{p(\bu_{i})} \ \bSigma_{p(\bv_{\tau})},
    \quad \bCov_{p(\bbeta, \bu_{i})}, \ \bCov_{p(\bbeta, \bv_{\tau})}, \
    \bCov_{p(\bu_{i}, \bv_{\tau})}, \\[1ex]
    1 \le i \le m, \ 1 \le \tau \le t,
  \end{array}
\end{equation}
are computed in the E-step using equation (\ref{eqn:postMeanVar}). Note that
(\ref{eqn:betasubentries}) are the sub vectors in the the posterior mean $\bmu_{p(\btheta)}$
and (\ref{eqn:sigmasubblocks}) are sub-block entries in the posterior variance
covariance matrix $\bSigma_{p(\btheta)}$. The na\"{i}ve EM algorithm is given in
Algortihm \ref{alg:EMforEB}.
%
\begin{algorithm}[h]
  \begin{center}
    \begin{minipage}[t]{150mm}
    \begin{small}
      \textbf{Initialize:} $\hat{\bSigma}^{(0)}$ \\[1ex]
      Set $\ell = 0$ \\[1ex]
      \textbf{repeat}
      \begin{itemize}
        \itemsep0em
         \item[] \textbf{E-step:} Compute $\bmu_{p(\btheta)}$ and
                 $\bSigma_{p(\btheta)}$ via equation (\ref{eqn:postMeanVar})
                 to obtain necessary mean and variance-covariance components needed
                 for the M-step.
         \item[] \textbf{M-step:} Compute variance components in $\hat{\bSigma}^{(\ell+1)}$
                 via equation (\ref{Eqn:EM_var_updates}).
         \item[] $\ell \leftarrow \ell + 1$
      \end{itemize}
      \textbf{until} log-likelihood converges
    \end{small}
    \end{minipage}
  \end{center}
  \caption{\textit{Na\"{i}ve EM Algorithm for empirical Bayes estimates
  of the variance components in the Bayesian mixed effects model as given in (\ref{MainMod1}) and (\ref{MainMod2}).}}
  \label{alg:EMforEB}
\end{algorithm}
%
%
The challenge in Algorithm \ref{alg:EMforEB} is computation of the
posterior mean vector $\bmu_{p(\btheta)}$ and posterior
variance-covariance matrix $\bSigma_{p(\btheta)}$ at each iteration.
We discuss the details of these challenges in Section \ref{Sec:CompCons}.
\subsubsection{Computational Challenges}\label{Sec:CompCons}
\begin{figure}[h]
   \centering
   \begin{subfigure}[t]{0.23\textwidth}
        \centering
        \includegraphics[height=1.5in]{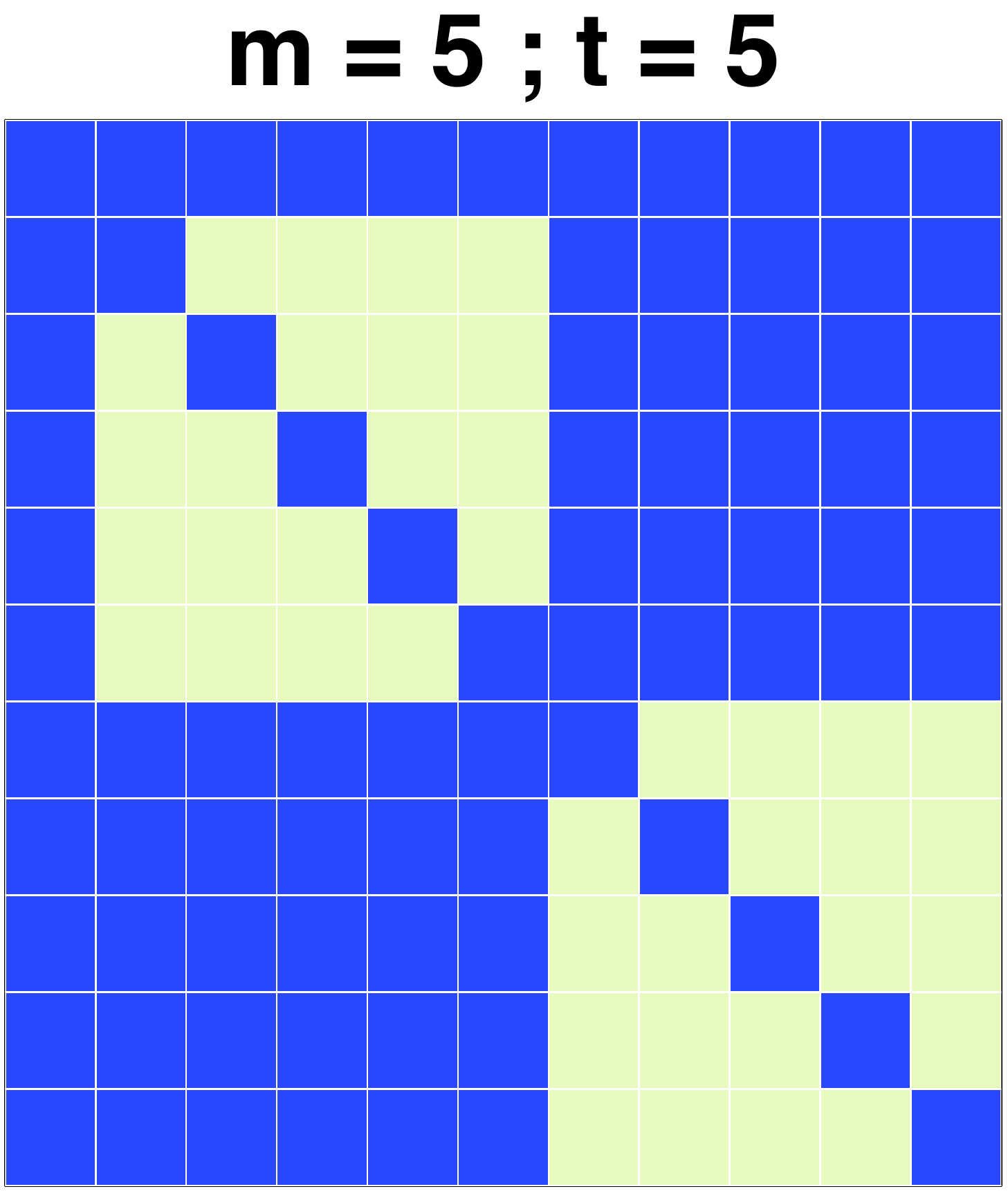}
    \end{subfigure}%
    \begin{subfigure}[t]{0.23\textwidth}
        \centering
        \includegraphics[height=1.5in]{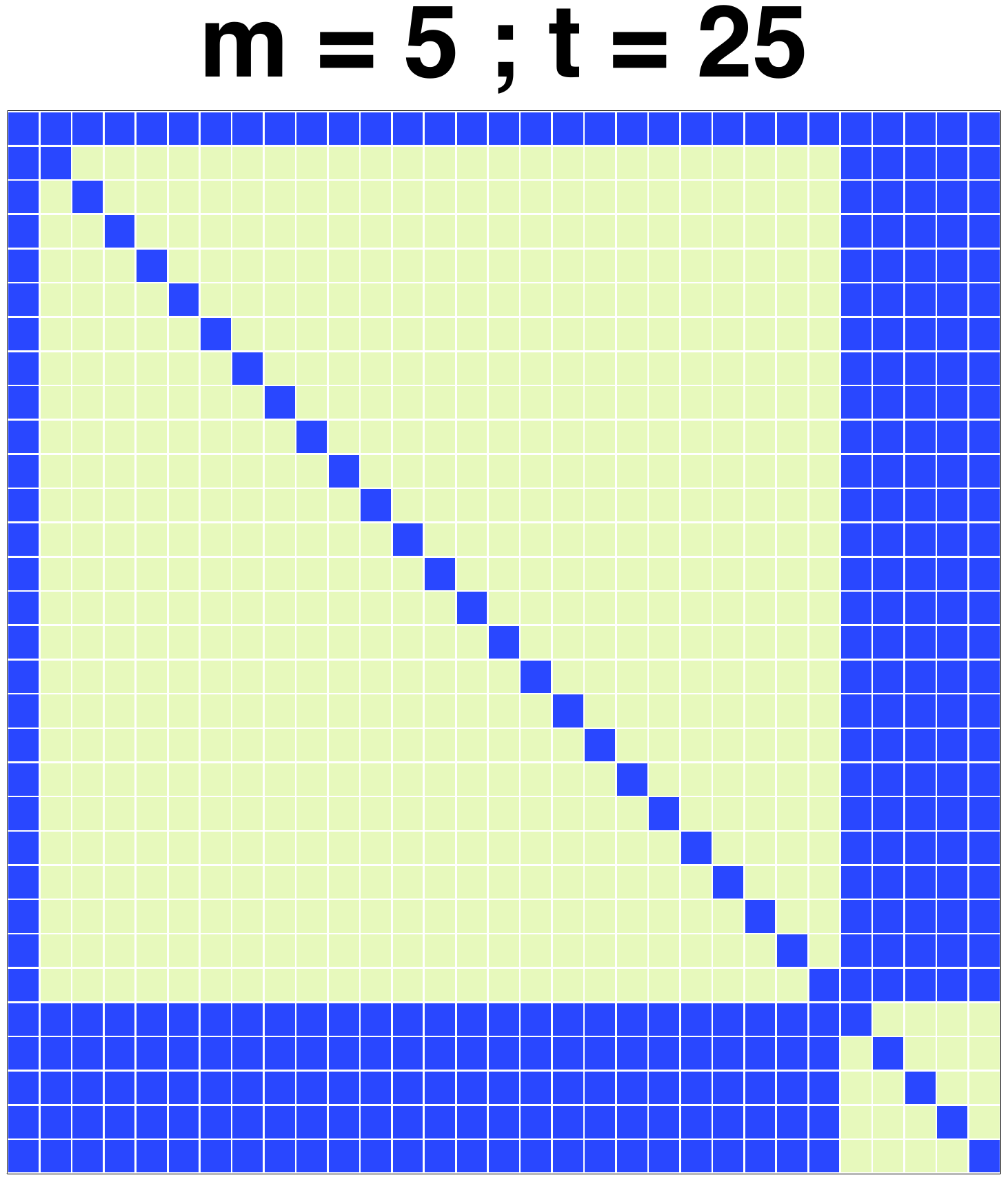}
    \end{subfigure}%
    \begin{subfigure}[t]{0.23\textwidth}
        \centering
        \includegraphics[height=1.5in]{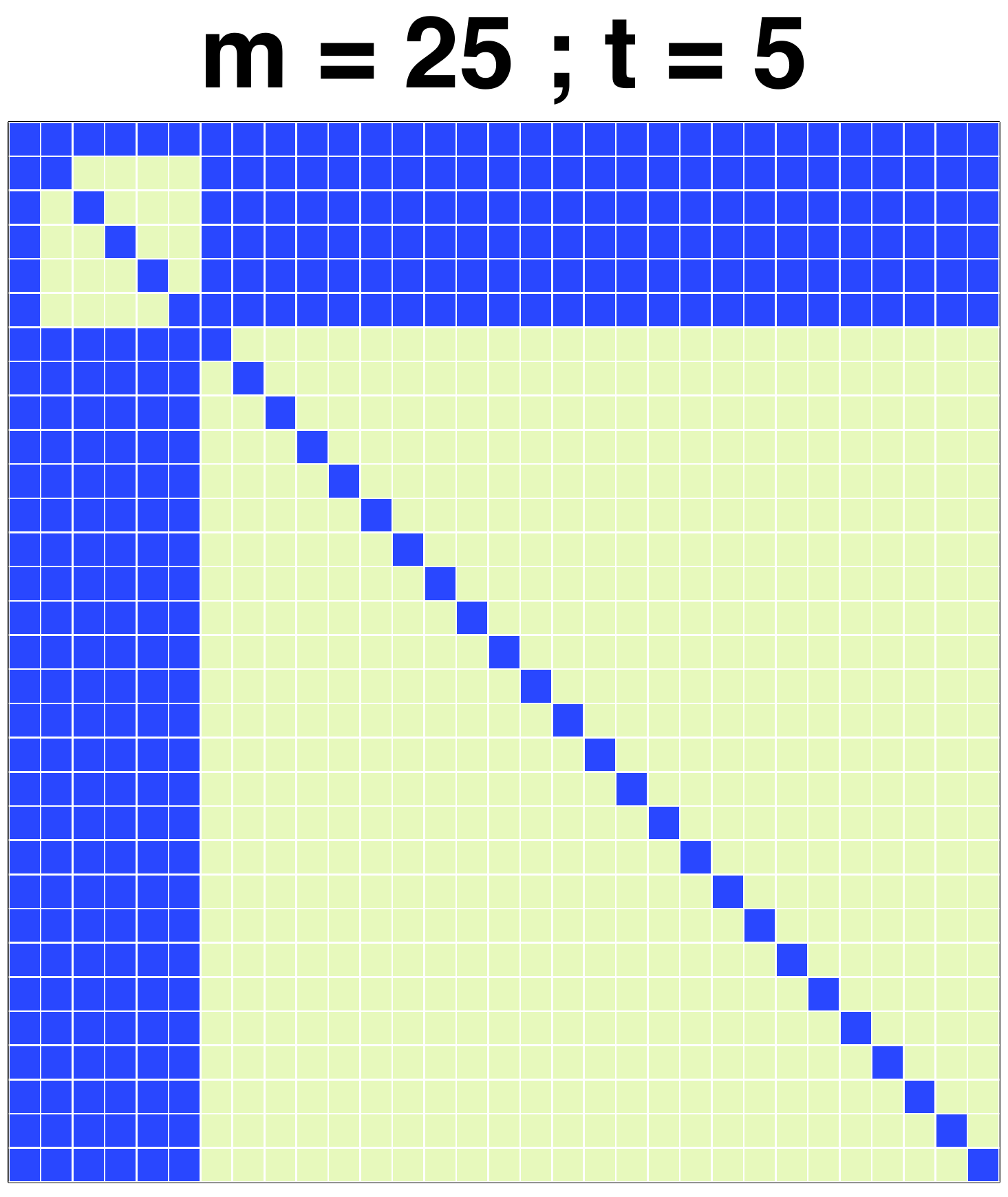}
    \end{subfigure}%
    \begin{subfigure}[t]{0.23\textwidth}
        \centering
        \includegraphics[height=1.5in]{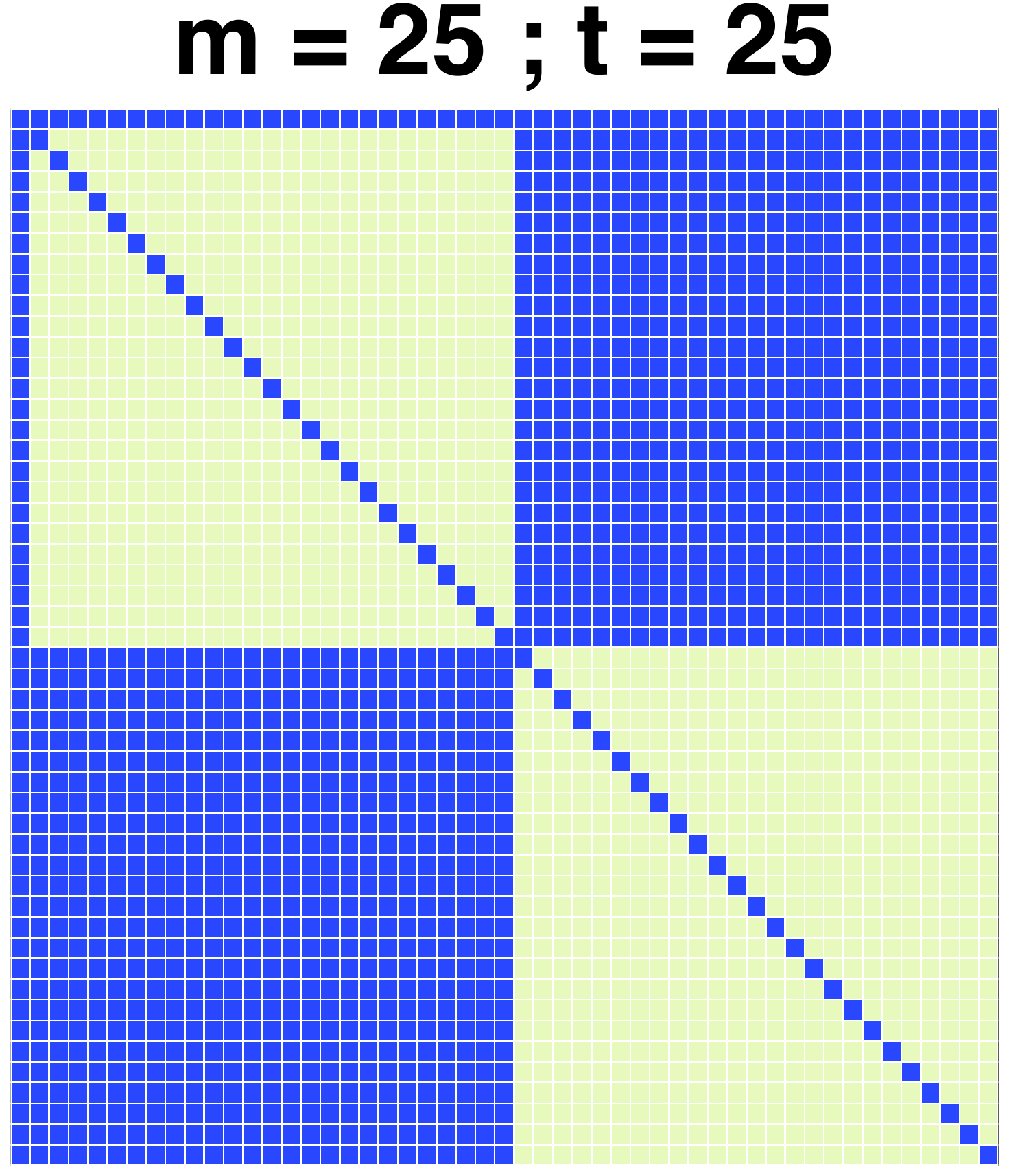}
    \end{subfigure}%
\caption{\textit{Sparsity present in $\bC^{\top} \bR^{-1} \bC + \bD$
                  under the Bayesian mixed effects model as represented
                  in (\ref{MainMod1}) and (\ref{MainMod2}). Here,
                  $p=q_{u}=q_{v}=1$. Non-zero $1 \times 1$ entries are
                  represented by a blue square and zero $1 \times 1$ entries
                  are represented by a light-yellow square.}}
  \label{fig:ModelIISparsity}
\end{figure}

At each iteration in Algorithm \ref{alg:EMforEB}, computation of
$p(\btheta)$ requires solving the sparse matrix linear system
\begin{equation}
  \label{eqn:sparseLinearSystem}
  \begin{array}{c}
    \bC^{\top} \bR^{-1} \bC + \bD = \bC^{\top} \bR^{-1} \bY + \bo
  \end{array}
\end{equation}
where the LHS of (\ref{eqn:sparseLinearSystem}) has sparse structure
 imposed by the random effects as exemplified in Figure \ref{fig:ModelIISparsity}.
%
%
This matrix has dimension
$$
  (p+m q_{u} + t q_{v}) \times (p+ m q_{u} + t q_{v}).
$$
It is often the case that the number of fixed effects parameters $p$,
the number of random effects parameters per user $q_{u}$ and
the number of random effects parameters per time $q_{v}$ are
of moderate size. Consequently, it is well known that
na\"{i}ve computation of $p(\btheta)$ is
$\mathcal{O}((m+t)^{3})$, that is, cubic dependence on the
number of random effects group sizes $m$ and $t$.

To address this computational problem, we employ the fact that in this setting
the matrix requiring inversion is sufficiently block-diagonal that its sparsity
can be exploited. In addition, the closed form updates of the variance components
in (\ref{Eqn:EM_var_updates}) require computation of only the sub-blocks
of $\bSigma_{p(\btheta)}$ that correspond to the non-zero sub-blocks of
$\bC^{\top} \bR^{-1} \bC + \bD$ (as illustrated in Figure \ref{fig:ModelIISparsity})
and not the entire matrix.
\subsection{Streamlined Empirical Bayes}\label{sec:streamPEB}
Streamlined updating of $\bmu_{p(\btheta)}$ and each of the sub-blocks of $\bSigma_{p(\btheta)}$
required for the E-step in Algorithm \ref{alg:EMforEB} can be embedded within the class
of two-level sparse matrix problems as defined in \cite{nolan2019streamlined} and is
encapsulated in Result 1. Result 1 is analogous and mathematically identical to Result 2 in
\cite{menictas2019streamlinedCrossed}. The difference being that the authors in
\cite{menictas2019streamlinedCrossed} do not apply their methodologies to the mobile health
setting and use full variational Bayesian inference for fitting as opposed to our use of
empirical Bayes.

%
%
\vskip2mm
\noindent
{{\bf Result 1} \textit{(Analogous and mathematically identical to Result 2 in
\cite{menictas2019streamlinedCrossed}).}} \textit{
The posterior updates under the Bayesian mixed effects model as given in
(\ref{MainMod1}) and (\ref{MainMod2}) for $\bmu_{p(\btheta)}$
and each of the sub-blocks of $\bSigma_{p(\btheta)}$ are expressible as
a two-level sparse matrix least squares problem of the form $|| \bb - \bB \
\bmu_{p(\btheta)} ||^{2}$ where $\bb$ and the non-zero sub-blocks of $\bB$,
according to the notation in the appendix, are, for $1 \le i \le m$,}
$$
  \bb_{i} \equiv \left[
    \begin{array}{c}
      \sigma_{\varepsilon}^{-1} \bY_{i} \\[1ex]
      m^{-\smhalf} \bSigma_{\bbeta}^{-\smhalf} \bmu_{\bbeta} \\[1ex]
      \bzero \\[1ex]
      \bzero
    \end{array} \right],
  \quad
  \bBdot_i \equiv \left[
    \begin{array}{c}
      \sigma_{\varepsilon}^{-1} \bZ^{\bu}_{i} \\[1ex]
      \bO \\[1ex]
      \bO \\[1ex]
      \bSigma_{u}^{-\smhalf}
  \end{array} \right],
  \quad
  \bB_{i} \equiv \left[
    \begin{array}{cc}
      \sigma_{\varepsilon}^{-1} \bX_{i} & \sigma_{\varepsilon}^{-1} \bZ^{\bv}_{i} \\[1ex]
      m^{-\smhalf} \bSigma_{\bbeta}^{-\smhalf} & \bO  \\[1ex]
      \bO & m^{-\smhalf} \left( \bI_{t} \otimes \bSigma_{\bv}^{-\smhalf} \right) \\[1ex]
      \bO & \bO
  \end{array} \right],
$$
\textit{with each of these matrices having $\nadj = t + p + t q_{v} + q_{u}$
rows. The solutions are
$$
  \bmu_{p(\bbeta)} = \mbox{ first p rows of } \bx_{1}, \quad
  \bSigma_{p(\bbeta)} = \mbox{ top left } p \times p \mbox{ sub-block of } \bA^{11},
$$
\begin{equation*}
  \begin{array}{l}
    \displaystyle{\stack{1\le i\le m}} \left( \bmu_{p(\bu_{i})} \right)
     = \mbox{subsequent $q_{u}\times 1$ entries of}\ \bx_{1}
     \mbox{following $\bmu_{p(\bbeta)}$},
  \end{array}
\end{equation*}
\begin{equation*}
  \begin{array}{l}
    \bSigma_{p(\bu_{i})} =\mbox{ subsequent $q_{u} \times q_{u}$ diagonal
    sub-blocks} \mbox{of $\AUoo$ following $\bSigma_{p(\bbeta)}$},
  \end{array}
\end{equation*}
\begin{equation*}
  \begin{array}{l}
    \bCov_{p(\bbeta, \bu_{i})} = \mbox{ subsequent $p \times q^{\prime}$
    sub-blocks of $\AUoo$} \mbox{to the right of $\bSigma_{p(\bbeta)}$},
        \ \ 1\le i \le m,
  \end{array}
\end{equation*}
\begin{equation*}
  \begin{array}{l}
    \bmu_{p(\bv_{\tau})}= \bx_{2,\tau}, \ \ \bSigma_{p(\bv_{\tau})} = \bA^{22,\tau}, \quad
    \bCov_{p(\bbeta, \bv_{\tau})} = \mbox{first $p$ rows of $\bA^{12,\tau}$}
  \end{array}
\end{equation*}
and
\begin{equation*}
  \begin{array}{l}
    {\displaystyle\stack{1\le i\le m}}
    \Big( \bCov_{p(\bu_{i}, \bv_{\tau})} \Big) =
    \mbox{ remaining $q_{u}$rows of $\bA^{12,\tau}$,}
  \end{array}
\end{equation*}
$1 \le \tau \le t$, where the $\bx_{1}$, $\bx_{2,\tau}$, $\bA^{11}$,
$\bA^{22, \tau}$ and $\bA^{12, \tau}$ notation is given in the appendix.}
%

%
The streamlined equivalent of Algorithm \ref{alg:EMforEB} is given in Algorithm
\ref{alg:EMforEBStream}. Algorihm \ref{alg:EMforEBStream} makes use of the
\SolveTwoLevelSparseLeastSquares\ algorithm which was first presented in
\cite{nolan2019streamlined} but also provided in the appendix of this article.
%
\begin{algorithm}[!th]
  \begin{center}
    \begin{minipage}[t]{150mm}
      \begin{small}
            \textbf{Initialize:} $\hat{\bSigma}^{(0)}$ \\[1ex]
            Set $\ell = 0$ \\[1ex]
            \textbf{repeat}
              \begin{itemize}
                 \item[] \textbf{E-step:} Compute components of $\bmu_{p(\btheta)}$ and sub-blocks of $\bSigma_{p(\btheta)}$:
                         \\[1ex]
                         $\begin{array}{c}
                           \Ssc\thickarrow \SolveTwoLevelSparseLeastSquares(\{(\bveci,\Bmati,
                           \Bmatdoti):1\le i\le m \})
                         \end{array}$
                         \\[1ex]
                         where
                        \\[1ex]
                        $\Ssc$ returns $\bx_{1}$, $\bA^{11}$, $\bx_{2, i}$,
                        $\bA^{22, i}$ and $\bA^{12, i}, 1\le i \le m$.
                 \item[] \textbf{M-step:} Compute variance components in $\hat{\bSigma}^{(\ell+1)}$ via equation (\ref{Eqn:EM_var_updates}).
                 \item[] $\ell \leftarrow \ell + 1$
              \end{itemize}
            \textbf{until} log-likelihood converges
      \end{small}
    \end{minipage}
  \end{center}
  \caption{\it Streamlined EM algorithm for empirical Bayes estimates of the variance
               components in the Bayesian mixed effects model as given in (\ref{MainMod1}) and (\ref{MainMod2}).}
  \label{alg:EMforEBStream}
\end{algorithm}
%
The computing time and storage for the streamlined updating of $\bmu_{p(\btheta)}$ and each of the sub-blocks
of $\bSigma_{p(\btheta)}$ required for the E-step in Algorithm \ref{alg:EMforEBStream}
becomes $\mathcal{O}(m t^3)$. For moderate sized $t$, this reduces to
$\mathcal{O}(m)$. If both $m$ and $t$ are large, one may resort to coupling the streamlining
present in this article with an approximation of the posterior so as to further reduce
computation time and storage. However, care needs to be taken with the choice of approximation
 so as to incur as little degradation in accuracy as possible. As
explained in Section 3.1 of \cite{menictas2015variational}, mean field variational
Bayes approximations tend to be very accurate for Gaussian response models. However,
such high accuracy does not manifest in general. Ignoring important posterior
dependencies via mean field restrictions often lead to credible intervals being too
small (e.g. \cite{wang2005inadequacy}).
\section{Performance Assessment and Comparison}\label{Sec:PerfAssComp}
In order to evaluate the speed achieved by our streamlined empirical Bayes algorithm,
we compare the timing and accuracy of our method against state of the art software,
\texttt{GPyTorch} \cite{gardner2018gpytorch},
which is a highly efficient implementation of Gaussian Process
Regression modeling, with GPU acceleration. Note that the Gaussian linear mixed effects
model as given in (\ref{MainMod1}) and (\ref{MainMod2}) is equivalent to a
Gaussian Process regression model
with a structured kernel matrix induced by the use of random effects. For ease of
notation, in the following, we use \SEB\ to refer to the streamlined empirical Bayes
algorithm, \CPU\ to refer to empirical Bayes fitting using \texttt{GPyTorch} with CPU and \GPU\ to
refer to empirical Bayes fitting using \texttt{GPyTorch} with GPU.
The \SEB\ and \CPU\ computations were conducted using an Intel Xeon CPU E5-2683.
The \GPU\ computations were conducted using an Nvidea Tesla V100-PCIE-32GB.
\subsection{Batch Speed Assessment}\label{Sec:SpeedAss}
%
We obtained timing results for simulated batch data according to versions of the
Bayesian mixed effects model as given in (\ref{MainMod1}) and (\ref{MainMod2})
and for which both the fixed effects and random effects had dimension two, corresponding
to random intercepts and slopes for a single continuous predictor which was generated
from the Uniform distribution on the unit interval. The true parameter values were
set to
\begin{equation*}
    \begin{array}{c}
        \bbeta_{\mbox{\tiny true}}=\left[
        \begin{array}{c}
        0.58 \\
        1.98
        \end{array}
        \right],
        \quad
        \bSigma^{\bu}_{\mbox{\tiny true}}=
        \left[
        \begin{array}{cc}
        0.32 & 0.09\\
        0.09 & 0.42
        \end{array}
        \right],
        \quad
        \bSigma^{\bv}_{\mbox{\tiny true}}=
        \left[
        \begin{array}{cc}
        0.30 & 0\\
        0 & 0.25
        \end{array}
        \right], \mbox{ and} \quad
        {\sigsqeps}_{\mbox{\tiny, true}} = 0.3,
    \end{array}
\end{equation*}
and, during the studies, the $t$ values were set to $30$, and the
number of datapoints specific to each user and time period, $n$,
was set to 5.
Four separate studies were run with differing
values for the number of users $m \in \{ 10, 50, 100,10000 \}$. The total number of
data points is then $ntm$, that is, $\mbox{datapoints} \in \{ 1500, 7500, 15000, 1500000 \}$.
We then
simulated 50 replications of the data for each $m$ and recorded the
computational times for variance estimation from \CPU, \GPU\ and \SEB.
Algorithm 2 was implemented in \texttt{Fortran 77}. The EM iterations
were stopped once the absolute difference between successive expected
complete data log-likelihood values fell below $10^{-5}$.
The stopping criterion was the same for gPyTorch and additionally the
maximum number of iterations was set to 15.

\begin{figure}[h]
  \begin{center}
      \includegraphics[width=0.6\linewidth]{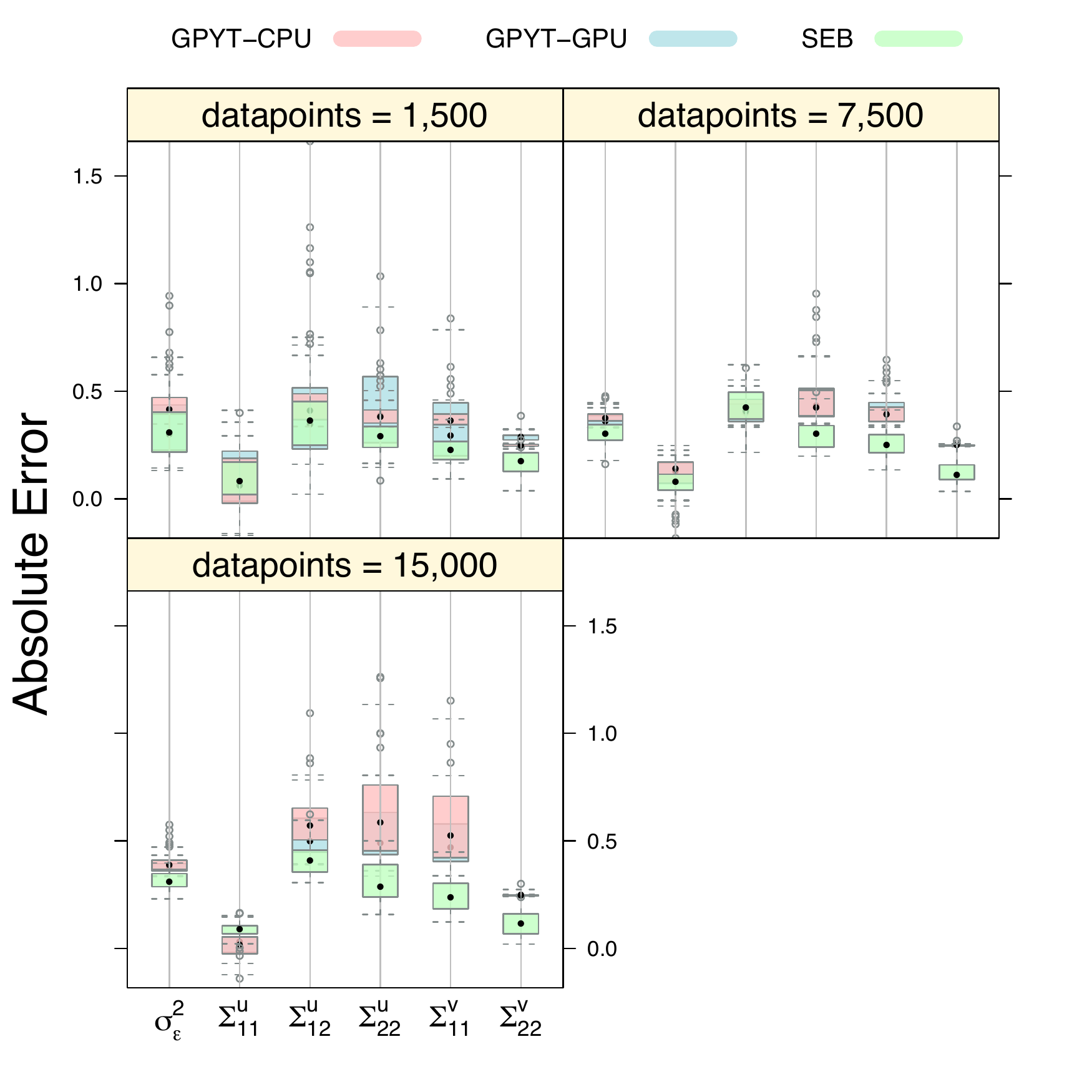}
      \caption{\textit{Summary of the simulation study in Section
      \ref{Sec:SpeedAss} under the Bayesian mixed effects model as
      represented in (\ref{MainMod1}) and (\ref{MainMod2}) where the
      absolute error values for each variance component estimated
      using one of three empirical Bayes methods summarized as a
      boxplot.}}
      \label{fig:AbsErrII}
    \end{center}
\end{figure}

Table \ref{tab:TimesII} shows the mean and standard deviation of elapsed
computing times in seconds for estimation of the variance components using
\SEB, \CPU\ and \GPU. Figure \ref{fig:AbsErrII} shows the absolute error
values for each variance components estimated using \SEB, \CPU\ and \GPU\
summarized as a boxplot.
%
\begin{table}[!ht]
\begin{center}
\scalebox{0.88}{
    \begin{tabular}{cccc}
    \hline\\[-2.3ex]
    Datapoints      & \SEB & \CPU & \GPU \\[0.2ex]
    \hline\\[-2.3ex]
    1,500     & 0.7 (0.10)    & 5.8 (0.14)    & 1.5 (0.16) \\[0.8ex]
    7,500     & 1.7 (0.15)    & 163.8 (1.81)  & 1.3 (0.04) \\[0.8ex]
    15,000    & 2.8 (0.21)    & 736.2 (38.36) & 5.2 (0.03) \\[0.8ex]
    1,500,000 & 322.1 (24.82) & NA (NA)       & NA (NA)    \\
    \hline
    \end{tabular}
}
\end{center}
\caption{\textit{Mean (standard deviation) of elapsed computing times in
seconds for estimation of the variance components in the Bayesian
  mixed effects model as represented in (\ref{MainMod1}) and (\ref{MainMod2}) using \SEB\ via Algorithm \ref{alg:EMforEBStream}, \CPU\ and \GPU\ for comparison.}}
\label{tab:TimesII}
\end{table}
%
%
%
\subsection{Online Thompson Sampling Contextual Bandit mHealth
Simulation Study}\label{sec:onlineAlg}
Next, we evaluate our approach
in a simulated mHealth study designed to capture many of the real-world
difficulties of mHealth clinical trials. Users in this simulated study
 are sent interventions multiple times each day according to Algorithm
\ref{alg:TSalgo}. Each intervention represents a message promoting
healthy behavior.

\begin{algorithm}[h]
  \begin{center}
    \begin{minipage}[t]{150mm}
    \begin{small}
      \textbf{Initialize:} $ \hat{\sigsqeps}, \ \hat{\bSigma}_{\bu}, \ \hat{\bSigma}_{\bv}$
      \\[1ex]
      \textbf{for $t \in \{ t_{1}, \hdots, t_{T} \}:$}
      \begin{itemize}
      \item[]
          \textbf{for $\tau = 1, \hdots, t:$}
          \begin{itemize}
            \itemsep0em
             \item[] Receive context features $\bX_{i\tau}$ for user $i$ and time $\tau$
             \item[] Obtain posterior $p(\btheta_{i \tau})$ using Result 1
             \item[] Calculate randomization probability $\pi$ in (\ref{eqn:randProb})
             \item[] Sample treatment $A_{i\tau} \sim \operatorname{Bern} \left({\pi}\right)$
             \item[] Observe reward $Y_{i \tau}$
          \end{itemize}
      \item[]
          \textbf{if $\tau = t:$}
          \begin{itemize}
            \itemsep0em
            \item[] Update hyper-parameters $\hat{\bSigma}$ with Algorithm \ref{alg:EMforEBStream}
            \item[] Update posterior $p(\btheta_{it})$
          \end{itemize}
    \end{itemize}
    \end{small}
    \end{minipage}
  \end{center}
  \caption{\textit{Thompson-Sampling algorithm with linear mixed effects model as given in
  (\ref{MainMod1}) and (\ref{MainMod2}) for the reward.}}
  \label{alg:TSalgo}
\end{algorithm}
%

In this setting there are
32 users and each user is in the study for 10 weeks. Users join the
study in a staggered fashion, such that each week new users might
be joining or leaving the study. Each day in the study users can
receive up to 5 mHealth interventions.

The Bayesian mixed effects model as represented in (\ref{MainMod1})
and (\ref{MainMod2}) offers several advantages in this setting.
In mHealth not only can users differ in the context that they
experience, but in their response to treatment under the same
context\cite{}. The user level random effects $\bu_{i}$ allow learning
of personalized policies for each user, overcoming the flaws of
methods which treat individuals as the same. Additionally, there can
be non-stationarity in how users respond to treatment, for example,
they might be more responsive in the beginning of a study than in the
end. By modeling time level random effects $\bv_{t}$, each person's
policy can be sensitive to a dynamic environment, and is informed
by how other users' responsivity has been influenced by time.

We evaluate our approach in a setting which demands personalization
within a dynamic environment. Users are modeled to be heterogenous
in their response to treatment. Additionally, their responsivity
declines with time.

\begin{figure}[h]
  \begin{center}
      \includegraphics[width=0.6\linewidth]{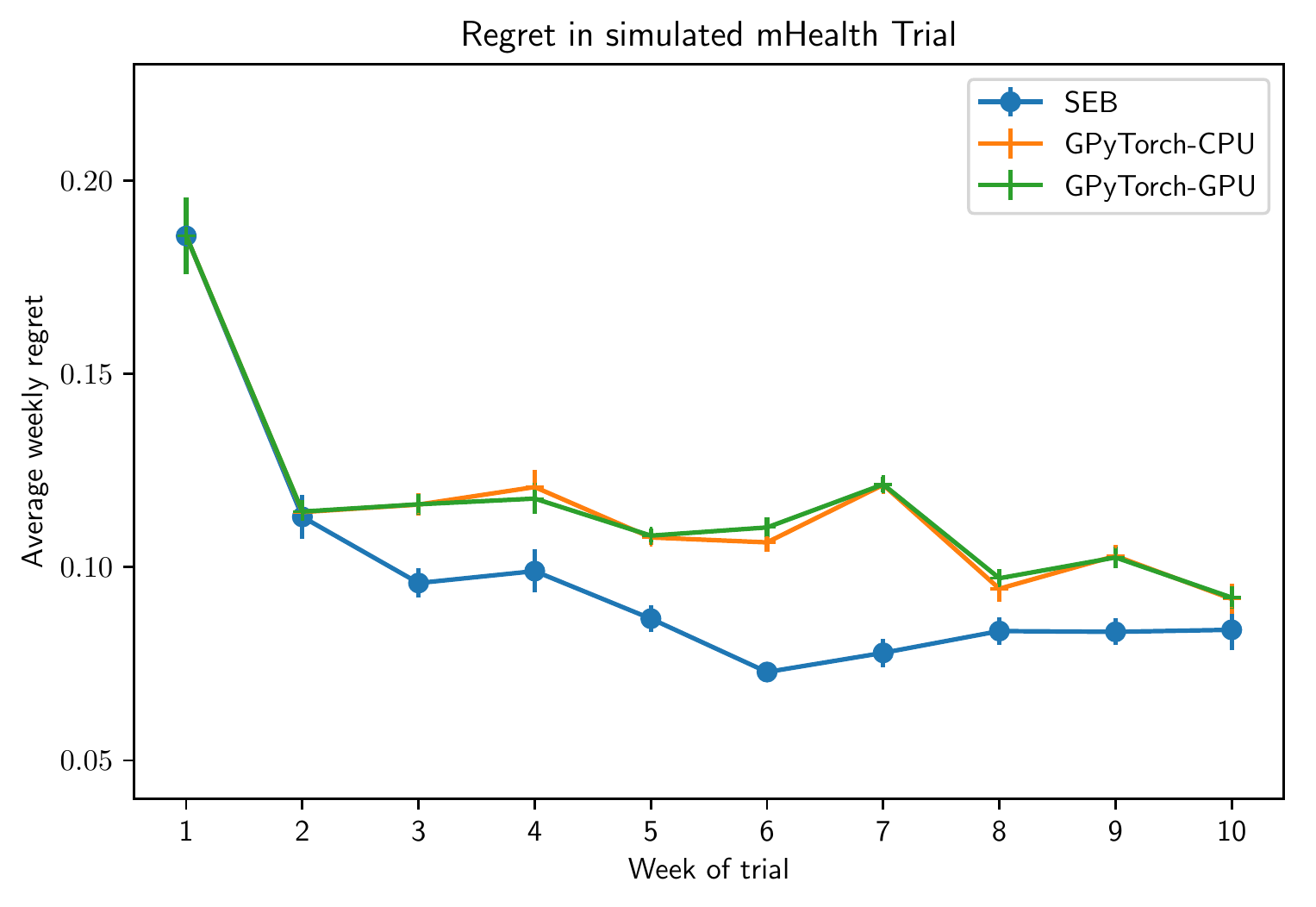}
      \caption{\textit{Regret averaged across all users for each week in the
      simulated mHealth trial associated with approaches \SEB, \CPU\ and \GPU.}}
      \label{fig:regretFig}
    \end{center}
\end{figure}

In Figure \ref{fig:regretFig} we show the ability of our streamlined
algorithm to minimize regret where real data is used to inform the simulation.
We also compare our approach to \CPU\ and \GPU. For all users we show
the average performance for their $n$th week in the study. For example,
we first show the average regret across all users in their first week
of the study, however this will not be the same calendar time week,
as users join in a staggered manner.
The average total time (standard deviation) for estimation of the
variance components was 757.1 (76.48) using \CPU, 7.5 (0.27) using
\GPU\ and 7.3 (0.16) using \SEB.

%
%
%
\section{Related Work}
The fundamental streamlined empirical Bayes Algorithm
\ref{alg:EMforEBStream} makes use of linear system solutions and sub-block
matrix inverses for two-level sparse matrix problems (\cite{nolan2019solutions}).
Our result for streamlined posterior computation in Section \ref{sec:streamPEB}
is analogous and mathematically identical to Result 2 in
\cite{menictas2019streamlinedCrossed} who instead focus on streamlined
mean field variational Bayes approximate inference for linear mixed models
with crossed random effects. In the present article, we make use
of a similar result for empirical Bayes posterior inference for use in the
mobile health setting. Our empirical Bayes algorithm allows streamlined estimation of
the variance components within an online Thompson sampling contextual bandit algorithm.

Other approaches include
using mixed model software packages for high-performance statistical
computation. For example: \emph{(i)} \texttt{BLME} provides a posteriori estimation
for linear and generalized linear
mixed effects models in a Bayesian setting \cite{dorie2015blme}; and \emph{(ii)}
\texttt{Stan} \cite{carpenter2017stan} provides full Bayesian statistical
inference with MCMC sampling. Even though \texttt{BLME} offers streamlined algorithms
for obtaining the predictions
of fixed and random effects in linear mixed models, the sub-blocks of the covariance matrices
of the posterior required for construction of the EM method in the streamlined empirical
Bayes algorithm are not provided by such software. On the other hand, \texttt{Stan} does
offer support for computation of these sub-blocks, but is well known to suffer
computationally in large data settings.

As we point to in Section \ref{Sec:PerfAssComp}, the Gaussian
linear mixed effects model used in the Thompson-Sampling
algorithm is equivalent to a Gaussian Process regression model
with a structured kernel matrix induced by the use of
random effects. Gaussian process models have been used for
multi-armed bandits
(\cite{chowdhury2017kernelized, hoffman2011portfolio, ambikasaran2015fast, srinivas2009gaussian, desautels2014parallelizing, wang2016optimization, djolonga2013high, bogunovic2016time}), and for contextual bandits (\cite{li2010contextual, krause2011contextual}).
To address the challenges
posed by mHealth, \cite{tomkins2019intelligent} illustrate the benefits of using
mixed effects Gaussian Process models in the context of reinforcement
learning. Computational challenges in the Gaussian
Process regression setting is a known and common problem
which has led to contributions from the computer
science and machine learning communities. For instance, to
address challenges posed for Gaussian Process regression
suffering from cubic complexity to data size, a variety of
scalable GPs have been presented, including the approach we compare to earlier:
\texttt{GPyTorch} \cite{gardner2018gpytorch}.
A review on state-of-the-art
scalable GPs involving two main categories: global approximations
which distillate the entire data and local approximations
which divide the data for subspace learning can be found in
\cite{liu2018gaussian}. The sparsity imposed by the use of random
effects, however, afford us accurate inference in the
cases considered in this article, and thus do not suffer from
the potential loss of accuracy that could result from the
approximate methods, such as those discussed in \cite{liu2018gaussian}.

\section{Discussion}
We compare three empirical Bayes approaches, \SEB, \CPU\ and \GPU\
for use within a batch simulation setting and an online contextual
bandit mHealth simulation study.

Within the batch simulation setting in Section \ref{Sec:SpeedAss},
inspection of the computational running times in Table
\ref{tab:TimesII} shows that \SEB\ achieves the lowest average
running time across all simulations and data set sizes, compared to
\CPU\ and \GPU. In the starkest case this results in a
98.96\% reduction in running time. Even in the more modest
comparison, \SEB\ timing is similar to that of \GPU\ but doesn't
require the sophisticated hardware that \GPU\ does.
The improvement between \CPU\ and \GPU\ is impressive and \GPU\
is clearly designed to excel in a resource rich environment.
However, in a clinical trial it is unclear if such an
environment will be available. In contrast, \SEB\
does not require advanced hardware to achieve state-of-the-art performance.

Our method makes use of computing only the necessary sub-blocks of the posterior
variance-covariance matrix at each time step, as opposed to computation
of the entire matrix. We were unable to template an equivalent streamlined
computation within \CPU\ and \GPU.
Consequently, we were unable to run \CPU\ and \GPU\ on the largest dataset
as this involves constructing a matrix of dimension
$(1.5 \times 10^6) \times (1.5 \times 10^6)$, even before the optimization procedure is called.
Templating the variance-covariance matrix such that it did not require this matrix as input
might have allowed us to run  \GPU\ on the largest dataset.
An advantage of our method is that it can
efficiently exploit the structure inherent within the  variance-covariance matrix
to manage large datasets.

The median reduction in error from \CPU\ to \SEB\ is 28.3\% and from
\GPU\ to \SEB\ is 22.2\%.
From Figure \ref{fig:AbsErrII}, we see that \SEB\ achieves lower
absolute error on average than either \CPU\ or \GPU. The
difference is more pronounced between \SEB\ and \GPU.
\SEB\ achieves the lowest average absolute error, at the fastest rate, on the
simplest machine.

\bibliographystyle{unsrt}
\bibliography{mybib}

\appendix
\section*{Appendix A.} \label{Sec:Appendix}

\subsection*{The \LargerSolveTwoLevelSparseLeastSquares\ Algorithm}\label{sec:STLSLS}

The \SolveTwoLevelSparseLeastSquares\ algorithm is listed in \cite{nolan2019solutions} and
based on Theorem 2 of \cite{nolan2019solutions}. Given its centrality to Algorithm \ref{alg:EMforEBStream}
we list it again here. The algorithm solves a sparse version of the the least squares problem:
$$\min_{\bx}\Vert\bb-\bB\bx\Vert^2$$
which has solution $\bx=\bA^{-1}\bB^T\bb$ where $\bA=\bB^T\bB$ and
where $\bB$ and $\bb$ have the following structure:
\begin{equation}
\bB\equiv
\left[
\arraycolsep=2.2pt\def\arraystretch{1.6}
\begin{array}{c|c|c|c|c}
\Bmato &\Bmatdoto &\bO &\cdots&\bO\\
\hline
\Bmatt &\bO &\Bmatdott&\cdots&\bO\\
\hline
\vdots &\vdots &\vdots &\ddots&\vdots\\
\hline
\Bmatm &\bO &\bO &\cdots &\Bmatdotm
\end{array}
\right]
\quad\mbox{and}\quad
\bb=\left[
\arraycolsep=2.2pt\def\arraystretch{1.6}
\begin{array}{c}
\bveco \\
\hline
\bvect \\
\hline
\vdots \\
\hline
\bvecm \\
\end{array}
\right].
\label{eq:BandbFormsReprise}
\end{equation}
The sub-vectors of $\bx$ and the sub-matrices of $\AtLev$ corresponding to its
non-zero blocks of are labelled as follows:
\begin{equation}
\bx=
\left[
\arraycolsep=2.2pt\def\arraystretch{1.6}
\begin{array}{c}
\bx_1\\
\hline
\bx_{2,1}\\
\hline
\bx_{2,2}\\
\hline
\vdots\\
\hline
\bx_{2,m}
\end{array}
\right]
\end{equation}
and
\begin{equation}
\AtLev^{-1}=
\left[
\arraycolsep=2.2pt\def\arraystretch{1.6}
\begin{array}{c|c|c|c|c}
\AUoo & \AUotCo & \AUotCt & \cdots  &\AUotCm \\
\hline
\AUotCoT & \AUttCo & \bigX & \cdots & \bigX \\
\hline
\AUotCtT & \bigX & \AUttCt & \cdots & \bigX \\
\hline
\vdots & \vdots  & \vdots & \ddots & \vdots \\
\hline
\AUotCmT & \bigX & \bigX & \cdots &\AUttCm \\
\end{array}
\right]
\label{eq:AtLevInv}
\end{equation}
with $\bigX$ denoting sub-blocks that are not of interest.
The \SolveTwoLevelSparseLeastSquares\ algorithm is given
in Algorithm \ref{alg:SolveTwoLevelSparseLeastSquares}.

\begin{algorithm}[h]
\begin{center}
\begin{minipage}[t]{150mm}
\begin{small}
\begin{itemize}
\setlength\itemsep{4pt}
\item[] Inputs: $\big\{\big(\bveci(\nadj_i\times1),
\ \Bmati(\nadj_i\times p),\ \Bmatdoti(\nadj_i\times q)\big): \ 1\le i\le m\big\}$
\item[] $\bomega_3\thickarrow\mbox{NULL}$\ \ \ ;\ \ \ $\bOmega_4\thickarrow\mbox{NULL}$
\item[] For $i=1,\ldots,m$:
\begin{itemize}
\setlength\itemsep{4pt}
\item[] Decompose $\Bmatdoti=\bQ_i\left[\begin{array}{c}
\bR_i\\
\bzero
\end{array}
\right]$ such that $\bQ_i^{-1}=\bQ_i^T$ and $\bR_i$ is upper-triangular.
\item[] $\cveczi\thickarrow\bQ_i^T\bveci\ \ \ ;\ \ \ \Cmatzi\thickarrow\bQ_i^T\Bmati$
        \ \ \ ; \ \ \ $\cvecoi\thickarrow\mbox{first $q$ rows of}\ \cveczi$
\item[] $\cvecti\thickarrow\mbox{remaining rows of}\ \cveczi$\ \ ;\ \
$\bomega_3\thickarrow
\left[
\begin{array}{c}
\bomega_3\\
\cvecti
\end{array}
\right]$
\item[]$\Cmatoi\thickarrow\mbox{first $q$ rows of}\ \Cmatzi$ \ \ \ ; \ \ \
       $\Cmatti\thickarrow\mbox{remaining rows of}\ \Cmatzi$ \ \ \ ; \ \ \
       $\bOmega_4\thickarrow
        \left[
        \begin{array}{c}
        \bOmega_4\\
        \Cmatti
        \end{array}
        \right]$
\end{itemize}
\item[] Decompose $\OmegaAtwoTwo=\bQ\left[\begin{array}{c}
\bR\\
\bzero
\end{array}
\right]$ such that $\bQ^{-1}=\bQ^T$ and $\bR$ is upper-triangular.
\item[] $\bc\thickarrow\mbox{first $p$ rows of $\bQ^T\bomega_3$}$
\ \ ; \ \ $\xveco\thickarrow\bR^{-1}\bc$ \ \ ; \ \
$\AUoo\thickarrow\bR^{-1}\bR^{-T}$
\item[] For $i=1,\ldots,m$:
\begin{itemize}
\setlength\itemsep{4pt}
\item[] $\xvectCi\thickarrow\bR_i^{-1}(\bc_{1i}-\Cmatoi\xveco)$ \ \ ; \ \
        $\AUotCi\thickarrow\,-\AUoo(\bR_i^{-1}\Cmatoi)^T$
\item[] $\AUttCi\thickarrow\bR_i^{-1}(\bR_i^{-T} - \Cmatoi\AUotCi)$
\end{itemize}
\item[] Output: $\Big(\xveco,\AUoo,\big\{\big(\xvectCi,\AUttCi,\AUotCi):\ 1\le i\le m\big\}\Big)$
\end{itemize}
\end{small}
\end{minipage}
\end{center}
\caption{\SolveTwoLevelSparseLeastSquares\ \textit{for solving the two-level sparse matrix
least squares problem: minimise $\Vert\bb-\bB\,\bx\Vert^2$ in $\bx$ and sub-blocks of $\bA^{-1}$
corresponding to the non-zero sub-blocks of $\bA=\bB^T\bB$. The sub-block notation is
given by (\ref{eq:BandbFormsReprise}) and (\ref{eq:AtLevInv}).}}
\label{alg:SolveTwoLevelSparseLeastSquares}
\end{algorithm}

\end{document}